\documentclass[10pt,twocolumn,letterpaper]{article}

\usepackage{multirow}
\usepackage{tabularx}
\usepackage[pagenumbers]{iccv} 

\definecolor{iccvblue}{rgb}{0.21,0.49,0.74}
\usepackage[pagebackref,breaklinks,colorlinks,allcolors=iccvblue]{hyperref}

\usepackage{marvosym}

\title{Integer Binary-Range Alignment Neuron for Spiking Neural Networks}

\author{Binghao Ye$^{1,7}$, Wenjuan Li$^{2}$, Dong Wang$^{1}$, Man Yao$^{4}$, Bing Li$^{2,5}$, Weiming Hu$^{2,3,6}$, \\ Dong Liang$^{1}$, Kun Shang$^{1}$\textsuperscript{\Letter} \\
$^1$ Shenzhen Institutes of Advanced Technology, CAS\\
$^2$ State Key Laboratory of Multimodal Artificial Intelligence Systems, CASIA \\
$^3$ School of Artificial Intelligence, University of Chinese Academy of Sciences \\
$^4$ Institute of Automation, Chinese Academy of Sciences  \\
$^5$ PeopleAI Inc. Beijing, China \\
$^6$ School of Information Science and Technology, ShanghaiTech University \\
$^7$ University of Chinese Academy of Sciences \\ 
bh.ye@siat.ac.cn , kunzzz.shang@gmail.com \\
}

\begin{document}
\maketitle
\begin{abstract}
Spiking Neural Networks (SNNs) are noted for their brain-like computation and energy efficiency, but their performance lags behind Artificial Neural Networks (ANNs) in tasks like image classification and object detection due to the limited representational capacity. To address this, we propose a novel spiking neuron, Integer Binary-Range Alignment Leaky Integrate-and-Fire to exponentially expand the information expression capacity of spiking neurons with only a slight energy increase. This is achieved through Integer Binary Leaky Integrate-and-Fire and range alignment strategy. The Integer Binary Leaky Integrate-and-Fire allows integer value activation during training and maintains spike-driven dynamics with binary conversion expands virtual timesteps during inference. The range alignment strategy is designed to solve the spike activation limitation problem where neurons fail to activate high integer values. Experiments show our method outperforms previous SNNs, achieving \textbf{74.19\%} accuracy on ImageNet and \textbf{66.2\%} mAP@50 and \textbf{49.1\%} mAP@50:95 on COCO, surpassing previous bests with the same architecture by +\textbf{3.45\%} and +\textbf{1.6\%} and +\textbf{1.8\%}, respectively. Notably, our SNNs match or exceed ANNs' performance with the same architecture, and the energy efficiency is improved by \textbf{6.3}${\times}$.

\end{abstract}    
\section{Introduction}
\label{sec:intro}

Artificial Neural Networks (ANNs) have gained prominence across various domains, including image recognition, object detection, and natural language processing, due to their remarkable capabilities~\cite{vaswani2017attention, ren2016faster, singhal2023large}. Nevertheless, their dependence on numerous multiply-accumulate operations (MACs) results in high energy consumption, presenting a substantial obstacle to widespread adoption, especially in resource-constrained environments such as edge computing. Conversely, Spiking Neural Networks (SNNs) offer an energy-efficient alternative by mimicking biological neural systems. A spiking neuron fires only when its membrane potential exceeds a threshold, producing 0/1 activation values: 1 (spike) or 0 (silence). This enables computational processes based on sparse accumulate operations (ACs), significantly reducing energy consumption on neuromorphic chips.

Unfortunately, the firing mechanism of SNNs often restricts the representational capacity of neurons, leading to significant flaws in both spatial representation and temporal dynamics~\cite{yao2025scaling}, which results in performance lags compared to ANNs in tasks like image classification and object detection. To address this, researchers have made numerous efforts. Real Spike~\cite{guo2022real} introduces a learnable coefficient to assign real values to the \{0, 1\} spike feature maps generated by spiking neurons. However, this method merely post-processes the spiking neuron outputs, akin to applying a non-shared convolution kernel, without fundamentally increasing information expression capacity. Ternary spikes~\cite{guo2024ternary} extend spiking neuron emissions to $\{-\alpha, 0, \alpha\}$, slightly enhancing the information expression capacity of the neuron but only marginally. The I-LIF~\cite{luo2024integer} neuron model extends the firing values of spiking neurons to any positive integer; however, its inference mode imposes a constraint on the maximum positive integer that a neuron can fire. These methods fail to fully address the issue of expression capability in neurons, causing SNNs to continue lagging behind ANNs in performance.

\begin{figure*}[ht]
\centering
\includegraphics[width=1.0\linewidth]{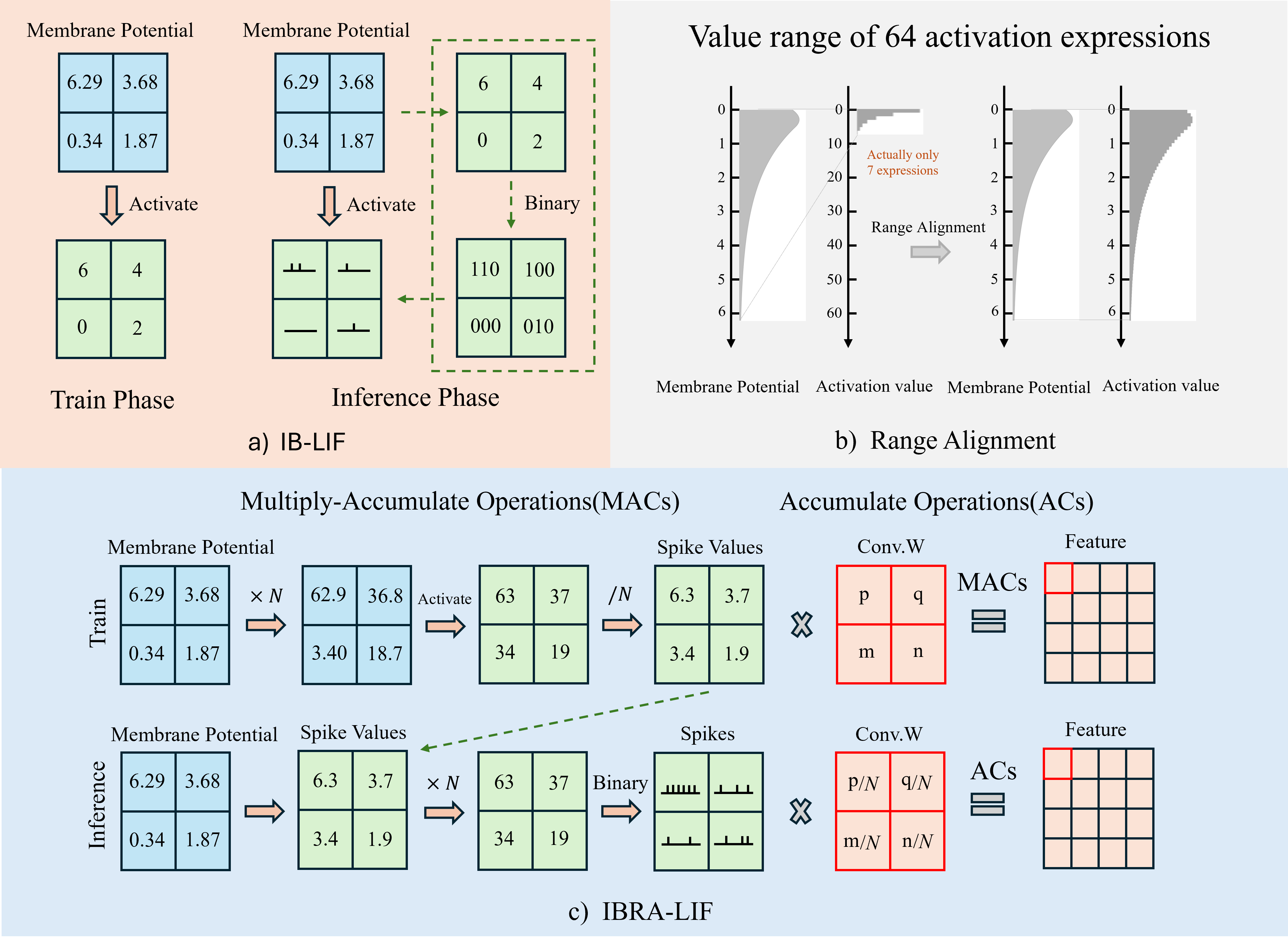}
\caption{ a) demonstrates the differences between IB-LIF during the training and inference phases. We implement binary conversion during the inference phase. b) illustrates the differences in activation value expressions before and after range alignment.  The theoretical value range for activation values encompasses 64 distinct expressions, however, before range alignment, only 7 expressions are actually represented. After applying range alignment, the complete range of activation values can be fully expressed. c) shows the overall workflow of N = 10 IBRA-LIF during different phases. During training, IBRA-LIF emits finer activation values to reduce information loss. During inference, IBRA-LIF converts the N-fold activation values to binary spikes and performs ACs with the re-parameterized convolution.}
\label{fig:IBRA-LIF}
\end{figure*}

In this work, we propose the Integer Binary-Range Alignment Leaky Integrate-and-Fire (IBRA-LIF) neuron for SNNs. IBRA-LIF extends the traditional spike encoding \{0, 1\} to a theoretically unlimited range, enabling more information expression. Importantly, the novel design comes with only a slight increase in energy consumption, ensuring the SNN models retain their characteristic energy efficiency.
Specifically, as depicted in \cref{fig:IBRA-LIF}, IBRA-LIF comprises an Integer Binary Leaky Integrate-and-Fire (IB-LIF) neuron and a range alignment (RA) strategy. IB-LIF enhances I-LIF by employing binary encoding, which addresses the limitations of I-LIF during inference. RA introduces a scaling factor $N$ to adjust the output range of spiking neurons, ensuring that the input and theoretical output ranges are properly aligned. Although RA introduces additional MACs, we implement a simple linear transformation to convert these MACs into ACs, and further reduce energy consumption through a re-parameterization technique, which preserves the key advantage of low energy consumption in SNNs. Additionally, we explore other possible strategies for spike activation limitation and demonstrate that our proposed RA strategy is optimal. We summarize our contributions as follows:
\begin{itemize}
\item[-] \textbf{IB-LIF:} By leveraging integer binary encoding, we propose the IB-LIF spiking neuron, which achieves an exponential increase in theoretical information expression capacity while maintaining low energy consumption.
\item[-] \textbf{Range alignment:} We introduce a range alignment strategy that effectively alleviates the spike activation limitation problem caused by expanding the maximum value of spike emissions. Furthermore, during inference, the MACs introduced by range alignment are transformed into ACs through a simple linear transformation, while using a re-parameterization technique, preserving the low energy consumption characteristic of SNNs.
\item[-] \textbf{Performance:} Our proposed method demonstrates outstanding performance in image classification and object detection tasks, achieving results on par with, or even surpassing, ANNs. Concretely, when using direct training SNN method, on ImageNet, it achieved $74.19\%$ accuracy, surpassing the prior state-of-the-art(SOTA) SNN by $3.45\%$. On COCO, it achieved $66.2\%$ mAP@50 and $49.1\%$ mAP@50:95, improving by $1.6\%$ and $1.8\%$ over the prior SOTA SNN with the same architecture. Moreover, we achieve performance that matches or exceeds ANN in the same architecture, with 6.3$\times$ energy efficiency. Importantly, we have also achieved near-lossless ANN-to-SNN conversion.
\item[-] \textbf{Versatility.} It is evident that our proposed IBRA-LIF can be extensively applied across various fields.  Its achievement of SOTA performance in image classification and object detection tasks heralds its potential value in numerous other domains.
\end{itemize}

\section{Related Work}
\label{sec:related}

\subsection{The Learning Algorithms of SNNs}
To achieve high-performance deep SNNs, two main learning strategies are currently employed: the ANN-to-SNN (A2S) conversion methods~\cite{cao2015spiking,rueckauer2016theory,hao2023reducing,sengupta2019going,ijcai2022p345} and the direct training (DT) methods~\cite{guo2023membrane,meng2023towards,wu2019direct}. A2S is to replace the activation functions in a pre-trained ANN with spiking neurons, aiming to approximate the activation values of ANN by using the average firing rates of SNN. However, the converted SNN typically requires a large number of timesteps to achieve accuracy comparable to that of the original ANN~\cite{han2020deep}, while neglecting the temporal dynamics of SNNs. DT employs surrogate gradient techniques to address the non-differentiability of spiking neurons, enabling the application of backpropagation. It effectively leverages the temporal dynamics of SNNs and allows for flexible architecture design, but training with multiple timesteps requires a significant amount of computational resources~\cite{yao2025scaling}. Our method achieves nearly lossless A2S with a small number of timesteps, and SOTA performance can still be achieved in DT with a small training timesteps.

\subsection{Information Loss and Expression in SNNs}
In SNNs, spiking neurons encode continuous inputs as 0/1 spikes, which inherently leads to information loss. To address this issue, several methods have been proposed~\cite{guo2023rmp,guo2022loss,guo2022real,guo2024ternary,luo2024integer}. For example, RMP-Loss~\cite{guo2023rmp} introduces a regularization term on the membrane potential, encouraging it to move closer to the spiking thresholds to reduce information loss. IM-Loss~\cite{guo2022loss} mitigates information loss by maximizing the information flow from the membrane potential to spikes. InfLoR-SNN~\cite{guo2022reducing} reduces the information loss by redistributing membrane potentials into a bimodal distribution near 0 and the spiking threshold. Despite these optimizations, traditional spiking neurons remain constrained by their 0/1 encoding, limiting their representational capacity. Methods, such as Real Spike~\cite{guo2022real}, Ternary Spike~\cite{guo2024ternary}, and I-LIF~\cite{luo2024integer}, aim to enhance the information expression capacity of spiking neurons through various strategies, but their improvements are marginal.
\section{Methodology}
\label{sec:method}
In this section, we present the Integer Binary-Range Alignment Leaky Integrate-and-Fire (IBRA-LIF) neuron, which is designed for SNNs. We start by outlining the necessary preliminaries. Next, we provide a thorough explanation of the operational principles and detailed mechanisms behind Integer Binary LIF (IB-LIF) neurons. Finally, we propose a specially designed range alignment (RA) strategy to tackle the spike activation limitation problem inherent in IB-LIF.

\subsection{Preliminaries}
\textbf{Leaky-Integrate-and-Fire (LIF) neuron:}\\
Leaky-Integrate-and-Fire(LIF) is currently the most favored neuron in SNNs, as it strikes a good balance between capturing the biological realism of neural systems and maintaining algorithmic simplicity.
Its simplified dynamics with soft reset can be described in three processes: charging, firing, and resetting. Concretely, the charging process is represented as:
\begin{equation}  \label{eq:lif1}
V^{t}_{pre} = \alpha V^{t-1} + I^t,
\end{equation}
where $V^t_{pre}$ represents the charged and pre-firing membrane potential of the spiking neuron at the $t$-th timestep, $\alpha$ is a time decay constant. $V^{t-1}$ is the membrane potential of the spiking neuron after resetting at the $(t-1)$-th timestep and $I^t$ is the input feature to the spiking neuron at the $t$-th timestep. When $V^t_{pre}$ exceeds a certain threshold, the neuron fires a spike, known as the firing process, which is expressed as follows:
\begin{equation}   \label{eq:lif2}
  O^t = \Theta (V^{t}_{pre} - V_{th})=
    \left\{
    \begin{aligned}
    1, {\quad}if{\quad}V^t_{pre}\geq{V_{th}}, \\
    0, {\quad}if{\quad}V^t_{pre}<{V_{th}}, \\
    \end{aligned}
    \right.
\end{equation}
where $\Theta$ is the step function, $V_{th}$ is a specific threshold, and $O^t$ represents the spike emitted by the spiking neuron at the $t$-th timestep, taking value of 1 if $x\geq 0$, and 0 otherwise.
Finally, the resetting process reduces the membrane potential of fired spike by $O^t$, as described by:
\begin{equation}   \label{eq:lif3}
  V^t = V^{t}_{pre} - O(t).
\end{equation}

Unfortunately, due to the simplicity of LIF’s output, the capacity for neurons to transmit and represent information is significantly constrained.\\
\textbf{Integer Leaky Integrate-and-Fire neuron:}\\
To address this challenge, the Integer Leaky Integrate-and-Fire (I-LIF) neuron~\cite{luo2024integer} modifies the spike emission mechanism of LIF, which allows neurons to emit a series of positive integers. The spiking output $O^t$ of I-LIF at the $t$-th timesteps is instead represented by:
\begin{equation}  \label{eq:I-lif1}
O^t = Clip(round(V^{t}_{pre}),0,D),
\end{equation}
where $round$ is a rounding function, $Clip(x, 0, D)$ means clipping $x$ to range $[0, D]$, and the hyperparameter $D$ determines the maximum integer value that I-LIF can output.
Notice that $O^t$ obtained from I-LIF by \cref{eq:I-lif1} is directly used for forward propagation in the training phase. During inference, I-LIF converts $O^t_l$ (where $l$ means the $l$-th layer) into a series of $\{0,1\}$ spikes $\{O^{t,d}_l\}_{d=1}^D$ and expands the timesteps $T$ to $T \times D$. Then, the following convolutional layer can be formulated as:
\begin{equation}  
Y^t_{l+1}=W_{l+1}\sum_{d=1}^DO_l^{t,d}=\sum_{d=1}^DW_{l+1}O_l^{t,d}
  \label{eq:I-lif2}
\end{equation}
where $Y^t_{l+1}$ is the output at the $t$-th timestep for the convolutional layer $l+1$, $W_{l+1}$ denotes the weight of convolutional layer $l+1$, and $O_l^{t,d}$ satisfies $\sum_{d=1}^DO_l^{t,d}=O^t_l$. 

I-LIF successfully extends the spike emission values from 0/1 to the positive integer, but when I-LIF encounters a large $D$, it requires more timesteps during inference, resulting in increased memory usage and energy consumption. As a result, I-LIF has not effectively addressed the challenge of enhancing the information expression capacity of spiking neurons.

\subsection{Integer Binary LIF}
To boost the information expression capacity of spiking neurons with low energy, we redesigned I-LIF to create Integer Binary LIF (IB-LIF), which uses binary integers during inference. This reduces energy exponentially and allows for an exponential increase in $D$ with similar energy levels. 

\begin{figure}[ht!]
\centering
\includegraphics[width=1.0\linewidth]{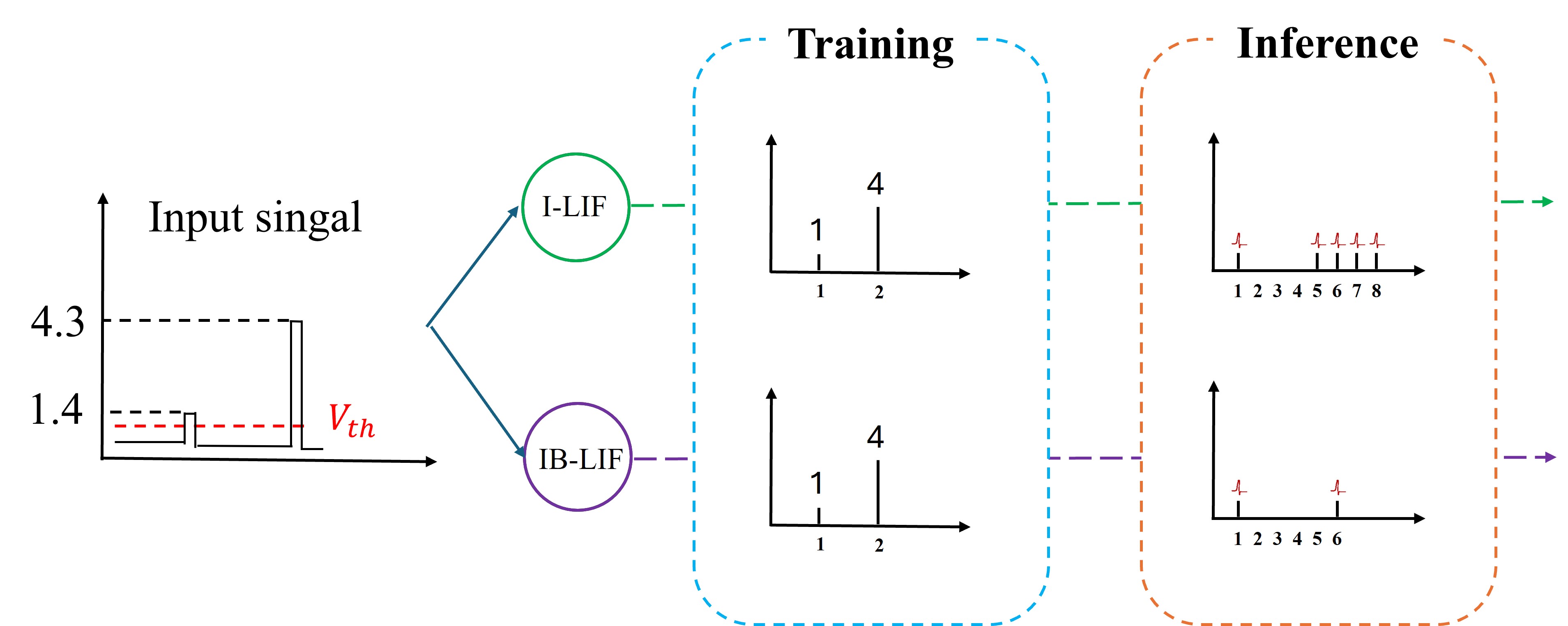}
\caption{The similarities and differences between IB-LIF and I-LIF. During training, IB-LIF emits positive integer values just like I-LIF. During inference, IB-LIF converts the positive integers into binary spikes, significantly reducing energy consumption. For example, I-LIF requires four spikes to represent the number 4, while IB-LIF only needs one.}

\label{fig:IB-LIF}
\end{figure}

As shown in \cref{fig:IB-LIF}, IB-LIF's output during training is identical to I-LIF's. However, during inference, IB-LIF is processed differently from I-LIF. Specifically, the neuron output $O^t_l$ during inference is changed as:
\begin{equation}\label{eq:IB-lif1}
  O^t_l = \sum_{b=1}^B 2^b O^{t,b}_l, where {\quad} O^{t,b}_l = (O^t_l>>b)\&1,
\end{equation}
where $B$ is the maximum number of bits required to convert $D$ into binary, “$>>b$” is right-shifting the binary representation of $O^t_l$ by $b$ positions, “$\&1$” represents performing a bitwise AND operation with 1, and $O^{t,b}_l$ represents the $b$-th bit of $O^t_l$ after conversion into binary. Then, \cref{{eq:I-lif2}} is rewritten as:
\begin{equation} \label{eq:IB-lif2}
 Y^t_{l+1}=W_{l+1}\sum_{b=1}^B 2^b O^{t,b}_l,
\end{equation}
Since convolution is a linear operator, we have:
\begin{equation} \label{eq:IB-lif3}
  W_{l+1}\sum_{b=1}^B 2^b O^{t,b}_l = \sum_{b=1}^B 2^b W_{l+1} O^{t,b}_l,
\end{equation}
Therefore, the output of the $l+1$-th convolutional layer during inference is formulated as:
\begin{equation}\label{eq:IB-lif4}
Y^t_{l+1}=\sum_{b=1}^B 2^b W_{l+1} O^{t,b}_l.
\end{equation}
Notice that $O^{t,b}_l$ in \cref{eq:IB-lif4} still comprises a sequence of 0/1 spikes. The computation involving $O^{t,b}_l$ and $W_{l+1}$ is carried out as ACs, thereby retaining the energy-efficient advantage of SNNs. Evidently, it allows us to exponentially scale $D$ while maintaining an energy consumption that is comparable to that of I-LIF (For example, increasing $D$ to 15 in IB-LIF has roughly equivalent energy consumption to when $D$ is 4 in I-LIF).

\subsection{Range Aligment}
\label{subsec:range aligment}
\begin{figure}[ht]
  \centering
   \includegraphics[width=0.8\linewidth]{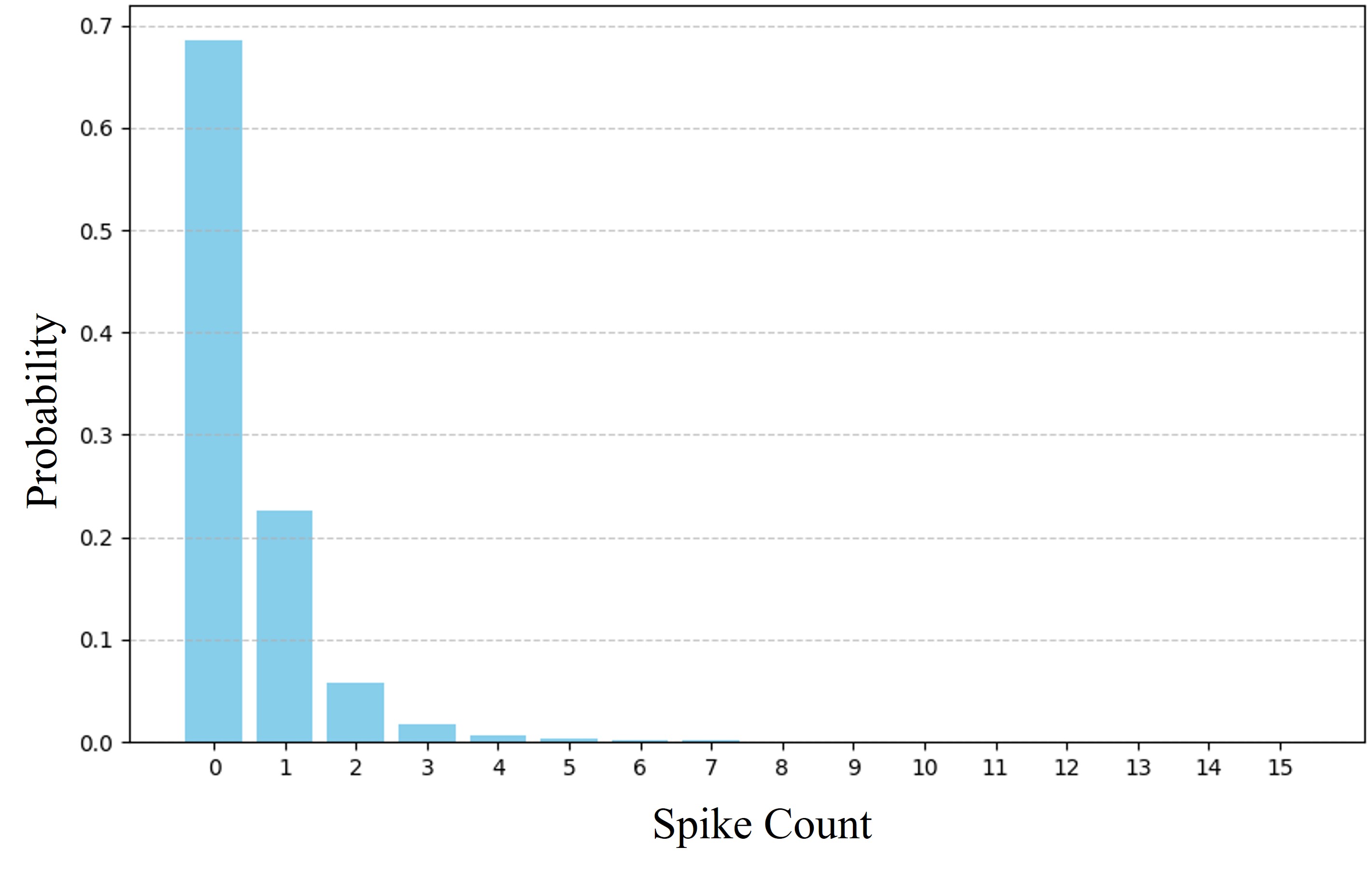}
   \caption{The spike activation limitation problem. when we set $D$ to 15, neurons fail to emit higher integer activation values, instead concentrating on lower ranges.}
   \label{fig:low-expression}
\end{figure}

IB-LIF can exponentially increase $D$ with a slight increase in energy consumption. However, attempting to significantly expand $D$ introduces a  
spike activation limitation problem. For instance, as illustrated in \cref{fig:low-expression}, the output range of IB-LIF fails to extend fully across the expected $[0, 15]$ range (Here $D=15$). Instead, the neuron output $O^t$ becomes overly concentrated within the lower numerical spectrum.  Consequently, the actual range $R_A^t=[0, 5]$ of $O^t$ is inconsistent with the theoretical range $R_T^t=[0, 15]$ at $t$-th timestep. This restricts the transmission of information to a narrow range of spike values, which we call the spike activation limitation problem, leading to insufficient information expression capacity in SNN models.

\begin{figure*}[t]
  \centering
   \includegraphics[width=1.0\linewidth]{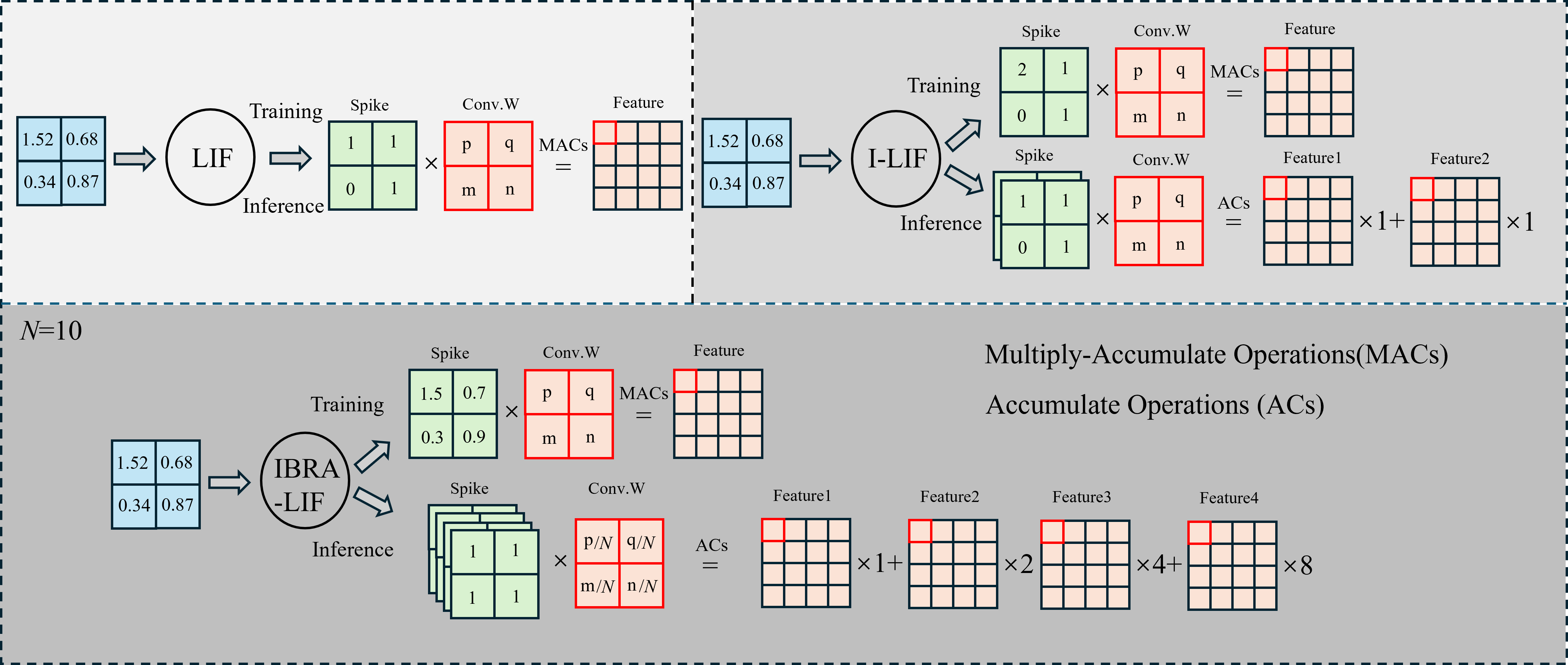}
   \caption{Comparison of LIF, I-LIF, and IBRA-LIF. LIF emits only 0/1 spikes during both training and inference, leading to significant errors. I-LIF emits positive integer spikes during training and converts them to 0/1 spikes during inference, but the number of ``1" spikes equals the integer value, which limits the information expression capacity of I-LIF. IBRA-LIF provides richer information representation through range alignment, and uses a simple linear transformation, re-parameterization technique, and binary conversion during inference, exponentially reducing the network's energy consumption. }
   \label{fig:range alignment}
\end{figure*}

To address it, we further introduce the range alignment (RA) strategy. RA involves setting a scaling factor $N$ to adjust the actual range $R_A^t$ and theoretical range $R_T^t$ of $O^t$, ensuring that it aligns with the range of $V^t_{pre}$. Specifically, we begin by multiplying $V^t_{pre}$ by $N$ to align its range with $R_T^t$, ensuring consistency among the range of $V^t_{pre} \times N$ , $R_A^t$ of $O^t$, and $R_T^t$ of $O^t$. After firing, we scale the $R_A^t$ and $R_T^t$ to match the range of $V^t_{pre} $. So, \cref{eq:I-lif1} is modified as: 
\begin{equation} \label{eq:IBRA-lif1}
    O^t = Clip(round(V^{t}_{pre}\times N),0, D_N)/N,
\end{equation}
Where $D_N$ is an integer that satisfies $D_N = D \times N$. However, at this point, $O_t$ is a floating-point number, which will introduce additional MACs, compromising the energy-efficient advantage of SNNs.

To ensure the energy efficiency of SNNs, we convert MACs to ACs through a simple linear transformation during inference. Specifically, the computation of the convolutional layer $(l+1)$ can be reformulated as:
\begin{equation} \label{eq:IR-lif3}
Y^t_{l+1} = W_{l+1}O^t_l = (W_{l+1}(O^t_l \times N))/N, 
\end{equation}
where $O^t_l \times N$ are integers, which can be expanded similar to \cref{eq:IB-lif1}:
\begin{equation} \label{eq:IR-lif4}
O^t_l\times N = \sum_{b=1}^B 2^b O^{N,t,b}_l,
O^{N,t,b} = ((O^t_l \times N)>>b)\&1,
\end{equation}
So, the output of the $l +1$-th convolutional layer of IBRA-LIF during inference is formulated as:
\begin{equation}
\label{eq:IR-lif5}
Y^t_{l+1} = (W_{l+1}(\sum_{b=1}^B 2^b O^{N,t,b}_l))/N,
\end{equation}
Finally, we can re-parameterize $W_{l+1}$ during inference to further reduce energy consumption. As shown in \cref{fig:range alignment}, \cref{eq:IR-lif5} is rewritten as:
\begin{align}\label{eq:IR-lif6}
Y^t_{l+1} &= W_{l+1}^N(\sum_{b=1}^B 2^b O^{N,t,b}_l)\\
          &= \sum_{b=1}^B 2^b W_{l+1}^N O^{N,t,b}_l.
\end{align}
where $W_{l+1}^N=W_{l+1}/N$, and $O^{N,t,b}_l$ is conprised of \{0,1\} spikes. 

Therefore, introducing RA enables IBRA-LIF to maintain the energy efficiency advantage of SNNs while overcoming the spike activation limitation problem in IB-LIF. 

Additionally, due to the non-differentiability of spiking neuron firing activity, we employ the Straight-Through Estimator surrogate gradients for direct training SNNs~\cite{Rathi2020Enabling}. The pseudo derivative for spike firing is as follows:
\begin{equation}
  \phi(V) =
    \left\{
    \begin{aligned}
    1 &, {\quad}if{\quad}0 \leq V \leq D \\
    0 &, {\quad} otherwise \\
    \end{aligned}
    \right.
  \label{eq:sg}
\end{equation}
where $D$ is the maximum value that IBRA-LIF can emit.

\subsection{Other possible strategies for spike activation limitation}
There are several straightforward strategies to solve the spike activation limitation problem, such as lowering the threshold of IB-LIF or addressing three major challenges: data normalization, batch normalization (BN) layers~\cite{ioffe2015batch}, and regularization. These strategies also help enhance the information expression capacity of the neuron's output $O^t$, causing the values of $O^t$ to spread across the entire range $[0, D]$. However, this may increase the risk of gradient explosion issues. Specifically, the forward propagation of the $l+1$-th convolutional layer is denoted as $Y^t_{l+1}=W_{l+1}O^t_l$, the backpropagation in SNNs can be formulated as:
\begin{equation}
  \frac{\partial L}{\partial W_{l+1}} = \sum_{t}\frac{\partial L}{\partial Y^t_{l+1}}\frac{\partial Y^t_{l+1}}{\partial W_{l+1}} = \sum_{t}\frac{\partial L}{\partial Y_{l+1}}O^t_l
  \label{eq:IR-lif2}
\end{equation}

Considering that $O^t_l$ is broadly distributed throughout the interval $[0, D]$, in this scenario, a high value of $O^t_l$, such as 15, could result in an excessively large model gradient. This, in turn, hinders the model from attaining stable convergence.

\section{Experiments} \label{sec:Exper}
In this section, we evaluate the effectiveness of IBRA-LIF for SNNs on two basic tasks, classification and detection. For the classification task, we have conducted experiments on three static image datasets, including CIFAR10~\cite{krizhevsky2009learning} dataset, CIFAR100~\cite{krizhevsky2009learning} dataset, and ImageNet~\cite{deng2009imagenet} dataset, and one event-based dataset, CIFAR10-DVS dataset~\cite{li2017cifar10}. For the detection tasks, we select the COCO val2017~\cite{lin2014microsoft} dataset as the benchmark.

\subsection{Experimental Settings}
Our experiments are on a 40-core Intel(R) Xeon(R) Gold 6242R 3.10GHz CPU and four NVIDIA RTX A6000 GPUs. The operating system is Ubuntu 20.04. In addition, we encode images into 0/1 spikes using the first layer of spiking neurons for static image classification tasks, as in most works~\cite{guo2022real,fang2021incorporating,guo2023joint}. For detection tasks, we utilize direct input encoding as in SpikeYOLO~\cite{luo2024integer}, repeating the image for $T$ timesteps as the input to the network. A detailed description of datasets and related experimental settings are included in the supplementary material. 

\subsection{Ablation Studies}
Hyperparameters $D$ (maximum value) and $N$ (scaling factor) significantly influence the IBRA-LIF neuron's performance. We performed the ablation studies to assess their effects, using ResNet18 as the backbone and the CIFAR10 dataset for experiments.

The selection criteria for hyperparameters $D$ and $N$ are strategically designed to maximize the benefits of binary representation and explore the impact of expression capacity at different magnitudes. Moreover, as indicated in \cref{subsec:range aligment}, excessively large values of $D$ are ineffective. We set $N = 10^n$, where $n \in {1, 2, 3}$, and choose $D$ to satisfy $D \times N = 2^B - 1$, with $D < 10$ (The choice of infinite $D$ is to validate the claims made in \cref{subsec:range aligment}). Note, in our binary-based method, the inference timestep is $T \times B$. The results are shown in \cref{tab:ab}. The optimal hyperparameters are $D =5.11$ and $N=100$, and it can be observed that when $D$ is unrestricted, the model’s performance declines, indicating that even allowing infinite values for $D$ does not yield positive returns. Unless stated otherwise, we set $N=100$ for subsequent experiments.

\subsection{Image Classfication}
We comprehensively compare IBRA-LIF with other methods on three popular datasets, CIFAR10 dataset, CIFAR100 dataset, and ImageNet dataset. The CIFAR10 results are presented in the supplementary materials, and the CIFAR100 and ImageNet results are reported in \cref{tab:cifar100,tab:imagenet}.

\begin{table}[ht!]
\caption{Ablation studies on CIFAR10 dataset. The maximum value of a neuron is allowed to emit as $D$, and the scaling factor for range alignment as $N$. $D \times N$ represent the information expression capacity. The timesteps $T$ all set to 1.}
\centering
\begin{tabular}{cccc}
\toprule
$D$ & $N$ & $D \times N$ & Acc  \\
\midrule
 1.5 & 10   & $2^4-1$   & 96.89\% \\
 3.1 & 10   & $2^5-1$      &96.94\% \\
 6.3 & 10   & $2^6-1$      &96.96\% \\
 1.27 & 100  & $2^7-1$     &96.76\% \\
 2.55 & 100  & $2^8-1$     &96.90\% \\  
 5.11 & 100  & $2^9-1$     &\textbf{97.10\%}\\
 $\infty$ & 100  & $\infty$ & 96.93\% \\  
 1.023 & 1000 & $2^{10}-1$    &96.50\% \\
 2.047 & 1000 & $2^{11}-1$    &96.93\% \\
 4.095 & 1000 & $2^{12}-1$    &96.78\% \\
 8.191 & 1000 & $2^{13}-1$    &97.09\% \\
 $\infty$ & 1000 & $\infty$ &96.83\% \\
\bottomrule
\end{tabular}
\label{tab:ab}
\end{table}

\begin{table}[ht!]
\caption{Performance of image classification on CIFAR100 dataset. In the previous SNNs, $D$ and $N$ are default to 1. }
\centering
\scalebox{0.8}{
\begin{tabular}{ccccc}
\toprule
Method & Type & Architecture & $T \times D$ & Acc  \\
\midrule
\multirow{2}{*}{ANN}  & \multirow{2}{*}{/} & ResNet18 & / & 80.12\%\\
                      &                               & ResNet19 & / & 81.56\% \\
\cline{1-5}
SlipReLU~\cite{jiang2023unified}          & A2S     & ResNet18 & 32 & 78.01\%\\
HT~\cite{Rathi2020Enabling}     & Hybrid  & VGG11 & 125 & 67.90\%\\
Real Spike~\cite{guo2022real}             & DT      & ResNet20 & 5 & 66.60\%\\
SLTT~\cite{meng2023towards}               & DT      & ResNet18 & 6 & 74.38\%\\ 
DSR~\cite{meng2022training}               & DT      & ResNet18 & 20 & 78.50\%\\
GAC-SNN~\cite{qiu2024gated}               & DT      & ResNet18 & 4 & 79.83\%\\
SLT-TET~\cite{anumasa2024enhancing}       & DT      & ResNet19 & 6 & 74.87\%\\
BKDSNN~\cite{xu2024bkdsnn}                & DT      & ResNet19 & 4 & 74.95\%\\
TAB~\cite{jiang2024tab}                   & DT      & ResNet19 & 6 & 76.82\% \\ 
LM-H~\cite{hao2023progressive}            & DT      & ResNet19 & 4 & 80.31\% \\
\multirow{2}{*}{CKA-SNN~\cite{zhang2024enhancing}} & \multirow{2}{*}{DT} & ResNet20 & 4 & 72.86\% \\ 
                                                    &                     & ResNet19 & 2 & 78.79\%\\    

\multirow{2}{*}{TTSpike~\cite{guo2024ternary}} & \multirow{2}{*}{DT} & ResNet20 & 4 & 74.02\% \\
                                                                &                     & ResNet19 & 2 & 80.20\%\\
\cline{1-5}                                                            
\multirow{2}{*}{IBRA-LIF(Ours)} & \multirow{2}{*}{DT} & ResNet18 & $1 \times 5.11$ & \textbf{80.16\%}\\
                                 &                     & ResNet19 & $1 \times 5.11$ & \textbf{81.67\%}\\   
\bottomrule
\end{tabular}
}

\label{tab:cifar100}
\end{table}

On the CIFAR100 dataset, our method demonstrates remarkable accuracies of \textbf{80.16\%} and \textbf{81.67\%} when utilizing ResNet18 and ResNet19, respectively. Our method achieves the best performance in the field, surpassing the prior SNN SOTA method by +\textbf{1.47\%} accuracy. Moreover, our method's performance is better than ANNs, exceeding the vanilla ANN's accuracy by +\textbf{0.11\%} with ResNet19.

As depicted in \cref{tab:imagenet}, our method achieves overwhelming success on the ImageNet dataset. Notably, our method, employing the ResNet18 architecture, surpasses the other SNN models with ResNet34. When leveraging the ResNet34 architecture, our method achieves an accuracy of \textbf{74.19\%} and outperforms the prior SOTA SNN by a remarkable +\textbf{3.45\%}, and exceeds the performance of ANN by +\textbf{0.35\%} with the same architecture. Besides, IBRA-LIF is naturally compatible with the A2S method. Without requiring any additional operations, we achieve a near-lossless conversion, attaining an accuracy of \textbf{73.04\%}, which surpasses SNNC-AP~\cite{li2021free} with 64 timesteps by +\textbf{1.92\%}.

Collectively, these results demonstrate the outstanding performance of IBRA-LIF in handling complex image classification tasks.

\subsection{Event-based Image Classfication}
Our method markedly surpasses the prior SOTA methods on the neuromorphic CIFAR10-DVS dataset, as evidenced by the results in \cref{tab:cifar-dvs}. Specifically, it achieves an accuracy of \textbf{83.10\%} and \textbf{83.50\%} using ResNet18 with 4 and 10 timesteps, exceeding the prior SOTA SNN by +\textbf{3.10\%} and +\textbf{3.50\%}, respectively. These results highlight the evident benefits of IBRA-LIF in processing neuromorphic datasets, effectively preserving the temporal dynamics of SNNs.

 \begin{table}[t]
\caption{Performance of event-based classification on CIFAR10-DVS dataset. All the SNN models are trained directly (DT).}
\centering
\scalebox{0.9}
{
\begin{tabular}{cccc}
\toprule
Method & Architecture & $T \times D$ & Acc  \\
\midrule
 STBP-tdBN~\cite{zheng2021going}               & ResNet19 & 10 & 67.80\% \\
 LM-H~\cite{hao2023progressive}                & ResNet19 & 10 & 79.10\% \\
\multirow{2}{*}{Real Spike~\cite{guo2022real}} & ResNet20 & 10 & 78.00\% \\
                                               & ResNet19 & 10 & 72.85\% \\
\multirow{2}{*}{TTSpike~\cite{guo2024ternary}} & ResNet20 & 10 & 79.80\% \\
                                               & ResNet19 & 10 & 79.80\% \\
\multirow{2}{*}{\centering CKA-SNN~\cite{zhang2024enhancing}}  & ResNet20 & 10 & 78.50\% \\  
                                                               & ResNet19 & 10 & 80.00\% \\
\cline{1-4}
\multirow{2}{*}{\centering IBRA-LIF(Ours)} & \multirow{2}{*}{ResNet18} & $4 \times 5.11$ &                                                \textbf{83.10\%} \\
                                        & & $10 \times 5.11$ & \textbf{83.50\%} \\

\bottomrule
\end{tabular}
}
\label{tab:cifar-dvs}
\end{table}

\begin{table}[t]
\caption{The energy consumption analysis on CIFAR10 dataset.}
\centering
\begin{tabular}{ccccc}
\toprule
Activation  & $T \times D$ & N & Acc & Emergy  \\
\midrule
Relu     & / & / & 96.99\%         & 2.79mJ\\
LIF      &  $4 \times 1$ & 1   & 95.45\%       & 0.38mJ\\
I-LIF    &  $1 \times 4$ & 1    & 96.23\%       & 0.15mJ\\
IBRA-LIF &  $1 \times 1.5$ & 10 & 96.89\% & 0.24mJ\\
IBRA-LIF &  $1 \times 5.11$ & 100 & 97.10\% & 0.44mJ\\
\bottomrule
\end{tabular}

\label{tab:emergy}
\end{table}

\subsection{Object Detection}
On the COCO dataset, as depicted in \cref{tab:coco}, our method significantly outperforms other SNN methods, and surpasses a similarly-sized YOLOv5 model by \textbf{+1.7\%} in mAP@50:95. Concretely,
our method achieves \textbf{63.8\%} mAP@50 and \textbf{47.1\%} mAP@50:95 with 23.1M parameters, and \textbf{61.3\%} mAP@50 and \textbf{44.8\%} mAP@50:95 with 13.2M parameters, which is a +\textbf{2.1\%} and +\textbf{2.3\%} improvement over SpikeYOLO using I-LIF with an equivalent number of parameters, respectively. Furthermore, our IBRA-LIF model with 48.1M parameters even outperforms SpikeYOLO using I-LIF with 68.8M.  These outcomes underscore the versatility of our IBRA-LIF method across various domains.

\begin{table*}[ht]
\caption{Performance of image classification on ImageNet dataset. }
\centering
\scalebox{1}{
\begin{tabular}{ccccc}
\toprule
Method & Type &Architecture & $T \times D$ & Acc  \\
\midrule
 \multirow{2}{*}{\centering ANN} & \multirow{2}{*}{\centering /} & ResNet18 & / & 71.18\%\\
                                 &                                          & ResNet34 & / & 73.84\%\\ 
\cline{1-5}
 signGD~\cite{oh2024sign} & A2S & ResNet34 & 32 & 58.09\% \\
 RMP-SNN~\cite{han2020rmp}    & A2S & ResNet34 & 512 & 60.08\%\\
 SlipReLU~\cite{jiang2023unified} & A2S & ResNet34 & 32 & 66.61\% \\
 SRP~\cite{hao2023reducing} & A2S & ResNet34 & 32 & 68.40\% \\
 QCFS~\cite{buoptimal} & A2S & ResNet34 & 32 & 69.37\% \\
 SNNC-AP~\cite{li2021free} & A2S & ResNet34 & 64 & 71.12\% \\
 \cline{1-5}
 IBRA-LIF(Ours) & A2S & ResNet34 & $1 \times 5.11$ & \textbf{73.04}\% \\ 
 \midrule
 BKDSNN~\cite{xu2024bkdsnn}            & DT      & ResNet18 & 4   & 65.60\%\\                      
 DSR~\cite{meng2022training}           & DT      & ResNet18 & 50 & 67.74\%\\
 InfLoR-SNN~\cite{guo2022reducing}     & DT      & ResNet34 & 4 & 65.54\%\\ 
 SEW ResNet~\cite{fang2021deep}        & DT      & ResNet34 & 4 & 67.04\%\\
 TAB~\cite{jiang2024tab}               & DT      & ResNet34 & 4 & 67.78\% \\
 GAC-SNN~\cite{qiu2024gated}           & DT      & ResNet34 & 6 & 70.42\%\\    
 \multirow{2}{*}{\centering CKA-SNN~\cite{zhang2024enhancing}} & \multirow{2}{*}{\centering DT} & ResNet18 & 4 & 62.95\%\\
                                                               &                                & ResNet34 & 4 & 66.78\% \\
 \multirow{2}{*}{\centering TTSpike~\cite{guo2024ternary}} & \multirow{2}{*}{\centering DT} & ResNet18 & 4 & 67.68\%\\                                                                           &                                & ResNet34 & 4 & 70.74\% \\  
\cline{1-5}
    \multirow{2}{*}{\centering IBRA-LIF(Ours)} & \multirow{2}{*}{\centering DT} & ResNet18 & $1 \times 5.11$  & \textbf{71.08\%}\\
                                           &                                & ResNet34 & $1 \times 5.11$ & \textbf{74.19\%}\\ 
\bottomrule
\end{tabular}
}
\label{tab:imagenet}
\end{table*}

\begin{table*}[ht]
  \caption{Performance of Object Detection on COCO dataset. Notice that the backbone of Meta-SpikeFormer is based on YOLOv5 and the IBRA-LIF models use the same backbone as SpikeYOLO.}
  \centering

  \begin{tabular}{cccccc}
    \toprule
    Method & Type &Param(M) & $T \times D$ & mAP@50(\%)& mAP@50:95(\%)  \\
    \midrule
    DETR~\cite{carion2020end} & / &41.0 & / & 62.4 & 42.0\\
    YOLOv5~\cite{ultralytics2021yolov5} & / & 21.2 & / & 64.1 & 45.4\\
    \cline{1-6}
    Spiking-Yolo~\cite{kim2020spiking}   & A2S &10.2 & 3500  & - & 25.7\\
    Bayesian Optim~\cite{kim2020towards} & A2S & 10.2 & 5000  & - & 25.9\\
    Spike Calib~\cite{li2022spike}       & A2S & 17.1& 512 & 45.4& - \\
    EMS-YOLO~\cite{su2023deep}                     & DT      & 26.9& 4 & 50.1 & 30.1\\ 
    \multirow{2}{*}{\centering Meta-SpikeFormer~\cite{DBLP:conf/iclr/YaoHHXZ00L24}} & \multirow{2}{*}{\centering DT}& 16.8 & 1 & 45.0 & -\\ 
                                                    &                    & 16.8 & 4 & 50.3 & -\\
    
    \multirow{4}{*}{SpikeYOLO~\cite{luo2024integer}} & \multirow{4}{*}{DT} & 13.2 & $1 \times 4$ & 59.2 & 42.5\\
                                                     &                     & 23.1 & $1 \times 4$ & 62.3 & 45.5 \\
                                                     &                     & 
                                                     48.1 & $1 \times 4$ & 64.6 & 47.4 \\
                                                     &                     & 
                                                     68.8 & $1\times4$ & 66.2 & 48.9 \\
    \cline{1-6}
    \multirow{3}{*}{IBRA-LIF(Ours)}  & \multirow{3}{*}{DT} & 13.2 & $1 \times 5.11$ & 61.3 & 44.8\\
                                      &                     & 23.1 & $1 \times 5.11$ & 63.8 & 47.1 \\
                                      &                     & 48.1 & $1 \times 5.11$ & \textbf{66.2} & \textbf{49.2}\\
  \bottomrule
  \end{tabular}

\label{tab:coco}
\end{table*}

\subsection{Energy Consumption Analysis}
We evaluate energy consumption during inference for a range of SNN models, including LIF, I-LIF, and IBRA-LIF, using the ResNet18 architecture on the CIFAR10 dataset (Supplementary materials include calculations of the specific energy consumption for both ANNs and SNNs). Moreover, we use vanilla ANN as a benchmark for comparison.

According to the data presented in ~\cref{tab:emergy}, when $T \times D = 1 \times 5.11, N=100$, despite IBRA-LIF consuming nearly three times as much energy as I-LIF, it is on par with LIF and has \textbf{6.3}$\times$ the energy efficiency compared to ANN. When $T \times D = 1 \times 1.5, N=10$, IBRA-LIF's energy consumption is \textbf{63\%} of that of LIF, while achieving a performance improvement of +\textbf{1.44\%}, considering both performance and energy efficiency, IBRA-LIF stands out as an excellent choice for SNNs.

It is important to note that, due to the limitations of current hardware, IBRA-LIF can only be implemented on synchronous neuromorphic chips. We believe that IBRA-LIF will eventually be realized on asynchronous neuromorphic chips in the future, allowing our method to perform inference with only $T$ timesteps.

\section{Conclusion} \label{sec:conc}
In this work, we propose an Integer Binary-Range Alignment LIF (IBRA-LIF) neuron for SNNs, which significantly improves the information expression capability of the spiking neuron while maintaining the advantage of low energy consumption. It leverages binary conversion, enabling an exponential increase in the maximum integer value a spiking neuron can emit with only a minimal increase in energy consumption. Then, a range alignment strategy is employed to mitigate the spike activation limitation problem caused by increasing the maximum integer value emitted by spiking neurons and preserving the energy efficiency of SNNs through a simple linear transformation and re-parameterization technique. Our extensive experiments on two core tasks—classification and object detection—demonstrate that IBRA-LIF achieves superior performance compared to previous methods.

{
    \small
    \bibliographystyle{ieeenat_fullname}
    \bibliography{main}
}

\end{document}